\documentclass[conference]{IEEEtran}
\IEEEoverridecommandlockouts
\usepackage{cite}
\usepackage{amsmath,amssymb,amsfonts}
\usepackage{algorithmic}
\usepackage{graphicx}
\usepackage{textcomp}
\usepackage{xcolor}
\def\BibTeX{{\rm B\kern-.05em{\sc i\kern-.025em b}\kern-.08em
    T\kern-.1667em\lower.7ex\hbox{E}\kern-.125emX}}

\usepackage{graphicx} 
\usepackage{gensymb}
\usepackage[]{algorithm2e}
\usepackage{cleveref}
\usepackage{amsmath}
\usepackage{siunitx}
\usepackage{tikz}
\usepackage{listings}
\usepackage{url}
\usepackage{colortbl}
\usepackage{subcaption} 

\usepackage{booktabs, multicol, multirow}

\usepackage{hyperref}

\newcommand{\doi}[1]{{doi:~\href{http://doi.org/#1}{#1}}\rmFullStop}

\newcommand*{\rmFullStop}{\rmifnextchar{.}{}{}}

\makeatletter
\newcommand{\rmifnextchar}[3]{%
  \begingroup
  \ltx@LocToksA{\endgroup#2}%
  \ltx@LocToksB{\endgroup#3}%
  \ltx@ifnextchar{#1}{%
    \def\next{\the\ltx@LocToksA}%
    \afterassignment\next
    \let\scratch= %
  }{%
    \the\ltx@LocToksB
  }%
}
\makeatother

\begin{document}
\title{\LARGE \bf
Machine Vision in the Context of Robotics: A Systematic Literature Review }\author{Robert Kirschner$^{1}$, Daniel Ro{\ss}burg$^{1}$ and  Javad Ghofrani$^{1}$
\thanks{$^{1}$Daniel Ro{\ss}burg, Robert Kirschner, and Javad Ghofrani are  with the Faculty of Informatics / Mathematics, Dresden University of Applied Sciences, Germany  %
		{\tt\small d.rossburg@gmail.com, robert.kirschner@htw-dresden.de, javad.ghofrani@gmail.com}}%
}
\author{
\IEEEauthorblockN{Javad Ghofrani}
\IEEEauthorblockA{\textit{HTW Dresden University of Applied Sciences} \\
\textit{Faculty of Informatics / Mathematics}\\
Dresden, Germany \\
javad.ghofrani@gmail.com}
\and

\IEEEauthorblockN{Robert Kirschner}
\IEEEauthorblockA{\textit{HTW Dresden University of Applied Sciences} \\
\textit{Faculty of Informatics / Mathematics}\\
Dresden, Germany \\
robert.kirschner@htw-dresden.de}
\and
\IEEEauthorblockN{Daniel Ro{\ss}burg}
\IEEEauthorblockA{\textit{HTW Dresden University of Applied Sciences} \\
\textit{Faculty of Informatics / Mathematics}\\
Dresden, Germany \\
d.rossburg@gmail.com}
\and

\IEEEauthorblockN{Dirk Reichelt}
\IEEEauthorblockA{\textit{HTW Dresden University of Applied Sciences} \\
\textit{Faculty of Informatics / Mathematics}\\
Dresden, Germany \\
dirk.reichelt@htw-dresden.de}
\and
\IEEEauthorblockN{Tom Dimter}
\IEEEauthorblockA{\textit{HTW Dresden University of Applied Sciences} \\
\textit{Faculty of Electrical Engineering}\\
Dresden, Germany \\
tom.dimter@htw-dresden.de}
}

\maketitle

\thispagestyle{plain}
\pagestyle{plain}


\begin{abstract}Machine vision is critical to robotics due to a wide range of applications which rely on input from visual sensors such as autonomous mobile robots and smart production systems. To create the smart homes and systems of tomorrow, an overview about current challenges in the research field would be of use to identify further possible directions, created in a systematic and reproducible manner. In this work a systematic literature review was conducted  covering research from the last 10 years. We screened 172 papers from four databases and selected 52 relevant papers. While robustness and computation time were improved greatly, occlusion and lighting variance are still the biggest problems faced. From the number of recent publications, we conclude that the observed field is of relevance and interest to the research community. Further challenges arise in many areas of the field. 
\end{abstract}


\section{Introduction} \label{sec:introduction}
Robotics is a fast growing research field with its wide spread application in various areas such as smart production systems, cyber physical systems and smart homes. 
Robots are now able to inspect their environments and make independent reactions to changes, discrepancies  and unforeseen situations. Higher quality in production, lower rejection rates and reduced costs are the results because of lower human maintenance. Intelligent robots would not be possible without the capability to react to their environment by vision and other sensors. A future without intelligent industrial and personal robots is not imaginable. As this development shows no signs of slowing, it is difficult to capture all current trends and challenges. The aim of this study is to determine the current state of the art and possible research gaps in the field of machine vision in the context of robotics through a systematic literature review, proposed by Kitchenham et al.~\cite{kitchenham2007guidelines}. The advantages of a systematic literature review are its reproducibility and repeatability which are ensured by its systematic execution and strict documentation. The review will be performed by using a strict review protocol to achieve maximum replicability and minimal bias. If performed again by independent researchers, only small differences should show in the results due to newly published papers. 
The method shares the initial steps with the Systematic Mapping Study proposed by Petersen et al.~\cite{petersen2015guidelines, petersen2008systematic}, but factors in the quality of the papers and provides much greater detail.

\subsection{Related Work}\label{sec:related_work}
\begin{figure}
 \centering
  \includegraphics[width=1.5\columnwidth]{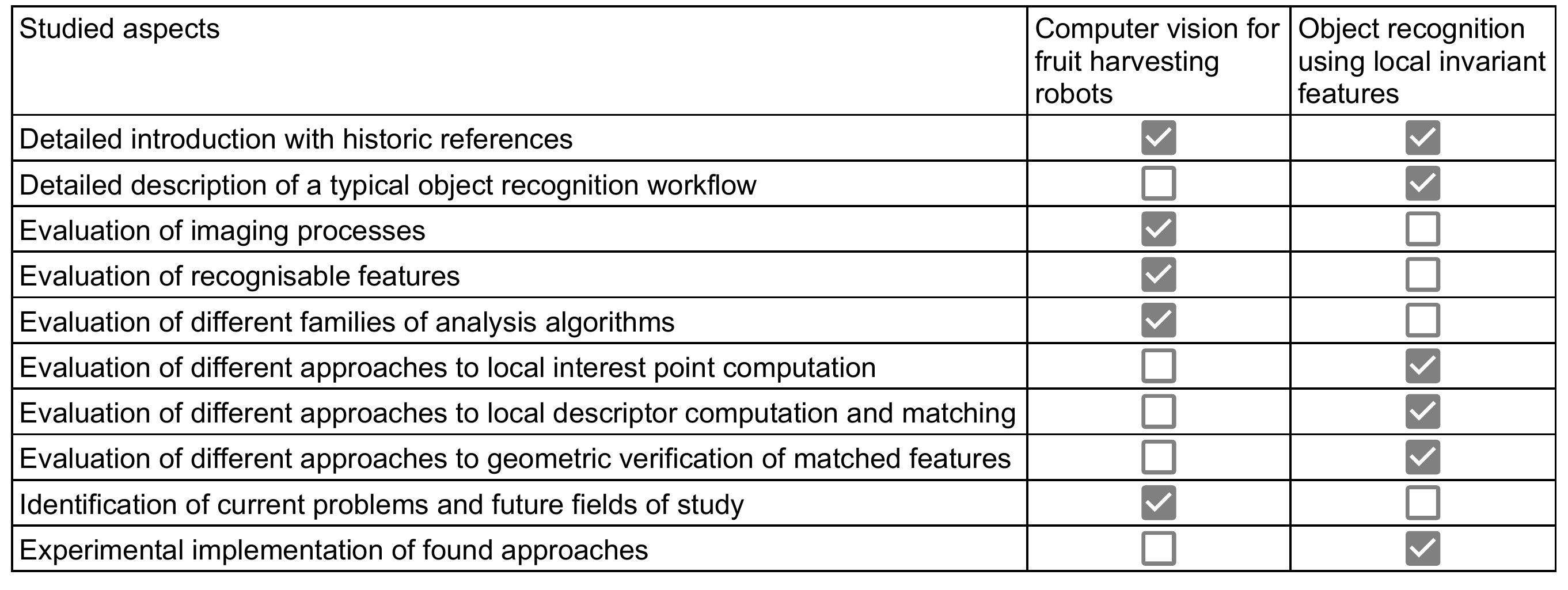}
  \caption{Comparison of two close related work}
  \label{fig:relatedTable}
\end{figure}

 Before carrying out our own literature research, it is important to look for related works that will tell us how other researchers have structured their evaluations and what aspects they have examined. Surprisingly few papers have been found that have similar objectives to our Literature Review. The first paper from 2012~\cite{kapach2012harvesting} is a review of the image recognition techniques that are or could be used in automated agriculture. Aspects of image recognition such as camera technology, recognizable features and recognition algorithms in the context of agricultural applications are discussed. Compared to industrial applications, the problems described are very similar, but sometimes even more challenging. Paper 2 from 2016~\cite{loncomilla2016invariant} promises to give an overview of the current methods for object recognition based on Local Invariant features. For each of the different stages of object recognition, the most important technologies are listed and described in detail. Finally, a research project based on the described technologies is carried out. Table \ref{fig:relatedTable} lists multiple facets, which have or haven't been part of theses two studies. The work of Kapach et al.~\cite{kapach2012harvesting} shows a broad overview over it's field, while~\cite{loncomilla2016invariant} shows a narrow slice of it in great detail. Furthermore an object tracking survey was conducted in~\cite{Liu2014}.
 Finally, Paper 3~\cite{Jena2017} reviews 19 neural network techniques used in image processing and its applications, weighting the pros and cons. Our approach is different in that it covers a broader field with no specific domain while following a strict review protocol. Additional material to our study is available to the interested reader on Figshare~\cite{figshare_ref_2019}.
\subsection{Objective}\label{sec:Objective}
This review will be performed by using a strict review protocol to achieve maximum repeatability and minimal bias. If performed again by independent researchers, only small differences should be visible in the results due to newly published papers. Therefore, if this study would be performed again later in time, the differences would show the progress of the research in this field of study (under the same conditions).

The rest of this paper is organized as follows: Section \ref{sec:methodology} describes the
methodology used to perform the literature review in detail, \ref{sec:results} presents the results and answers the research questions, finally \ref{sec:conclusion}  summarizes the work and gives an outlook.

\section{Research methodology} \label{sec:methodology}
\begin{figure}
  \centering
  \includegraphics[width=0.40\textwidth]{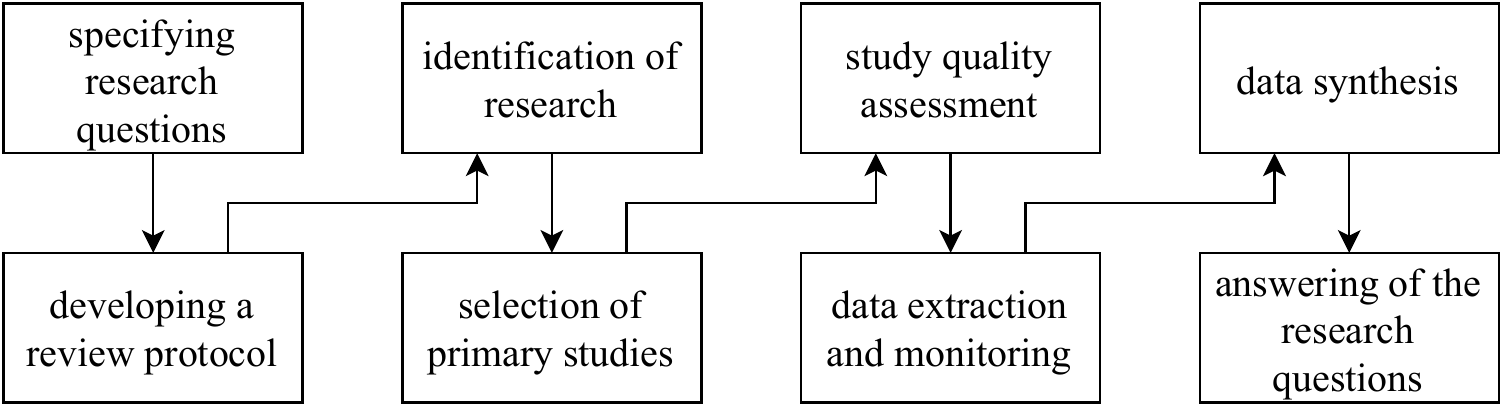}
  \caption{Procedure conducted for Systematic Literature Review}
  \label{fig:procedure}
  \end{figure}

The methodology used follows closely the approach proposed by~\cite{kitchenham2007guidelines} and shares the initial procedure steps of systematic mapping studies proposed by~\cite{petersen2008systematic}. 
First, the research questions for the review were defined, they should aim to produce the most relevant results in the chosen field of research. Next, a review protocol was created to specify and pin down all following steps. This allows the recreation of the study under the same conditions in the future.
The step selection of primary studies included building a search string and fine tuning it, followed by searching appropriate databases with it. Next, inclusion and exclusion criteria according to the research questions were applied to these results. By checking off a list of quality metrics, the quality of each paper was evaluated. Suitable metrics were  derived from the research questions to extract data from the data set. Afterwards the synthesizing of the acquired data by collating, summarizing and documenting the obtaining results was performed. Finally, the review was completed by answering the research question with the acquired data. The general steps of the procedure are displayed in Figure \ref{fig:procedure}.
 
 \subsection{Review Protocol}
The pre-defined protocol specifies the methods that were used to carry out the
systematic review. The protocol is necessary to reduce the possibility of researcher bias and improves replicability. 
 \subsection{Research Questions}
 The research questions to be addressed by this study are: 
 \begin{itemize}
     \item \textbf{RQ1: }Studies on which image processing techniques result in the most practical applications?
     \item \textbf{RQ2: }Which problems are typically solved by the studies in this research field and which ones are still unsolved?
    
 \end{itemize}
 \subsection{Search Process} \label{subsec:searchProcess}
 The search for papers was conducted via a predefined search string, which was adapted to each utilized research database. The aim was to collect around 150 initial papers, which were  later reduced to around 50 papers by applying inclusion and exclusion criteria. We limited the search area to 2009 - October 2018, because Nvidia began supporting GPU computation by providing the CUDA platform at that time~\cite{Shams2007} that allowed developers to directly utilize the power of GPUs for computational tasks. The used databases together with the respective search string and the number of results are shown in Table \ref{tab:researchTable}.
 \begin{table}[htbp]
\centering
\begin{tabular}{l|c|r}
\toprule
\textbf{Database} & \textbf{Searchstring} & \textbf{Number of Results}  \\\hline
IEEE & Figure \ref{fig:searchstringIEEE} & 40    \\ 
Scopus & / & 105    \\  
ACM & Figure \ref{fig:searchstringACM} & 9  \\
Web Of Knowledge & / & 18  \\ 
\bottomrule
\end{tabular}
\caption{used databases, search strings and number of results.}
\label{tab:researchTable}
\end{table}

 \subsection{Inclusion and exclusion of papers}
 Inclusions and exclusions are criteria which define if a paper that fits the search criteria is suitable for the final set. These were applied manually after reading the abstract and keywords for each paper. This process was used to further reduce the amount of papers and only keep the most relevant ones. Two researchers have completed this task independently to reduce subjectivity. Differences in decisions were discussed and resolved. The final set contained 52 papers. 
 
    \subsubsection*{Inclusion Criteria}
    papers and articles since 2009, from a computer science or engineering background which uses a deep learning approach, peer reviewed, experience reports, pose estimation systems, vision based gesture recognition, visual servoing, experiments
    \subsubsection*{Exclusion Criteria}
    non-English  articles  or  sources of  subjective  quality  like  summaries   or   keynotes,   keywords only  in  background  of  abstract, extreme  or  very  specialized  applications, secondary studies, no reference to the topic given

 \subsection{Quality Assessment} \label{subsec:QualityAssessment}
 Each papers quality was evaluated by a set of 14 predefined boolean questions of a specific area with an assigned numeric value. The value was modelled after the importance of each aspect for the quality of a paper. The higher the overall score of a paper, the higher is its apparent quality. The used quality measurements are presented in Table \ref{tab:qualityMeas}. The quality was measured by both researchers, then the mean value was calculated.

 \subsection{Data Collection}
 The answers to the following questions were extracted from each paper:
 \begin{itemize}
    \item The year when the paper was published
    \item Quality score for the study
    \item From which field of research does this study come? e.g. visual servoing, robot navigation, robot manipulation.
    \item How direct can the results be transferred into practical applications? (0 - purely theoretical treatise 10 - finished product was presented)
    \item On which existing methods is the approach based?
    \item Which improvements where made through the study?
    \item In which areas where improvements made?
    \item What is the magnitude of the progress made?
 \end{itemize}
 The data was extracted by both researchers, differences being resolved manually.
 
 \subsection{Data Analysis}
 The data was filled in a spreadsheet to show basic information and the collected answers for each paper. 
 
\section{Results} \label{sec:results}
  In this section, the collected data from the reviewed papers will be examined and analyzed. 
  \subsection{Meta data}
    \subsubsection{Publication Year}
\begin{figure}
\centering
  \includegraphics[width=0.40\textwidth]{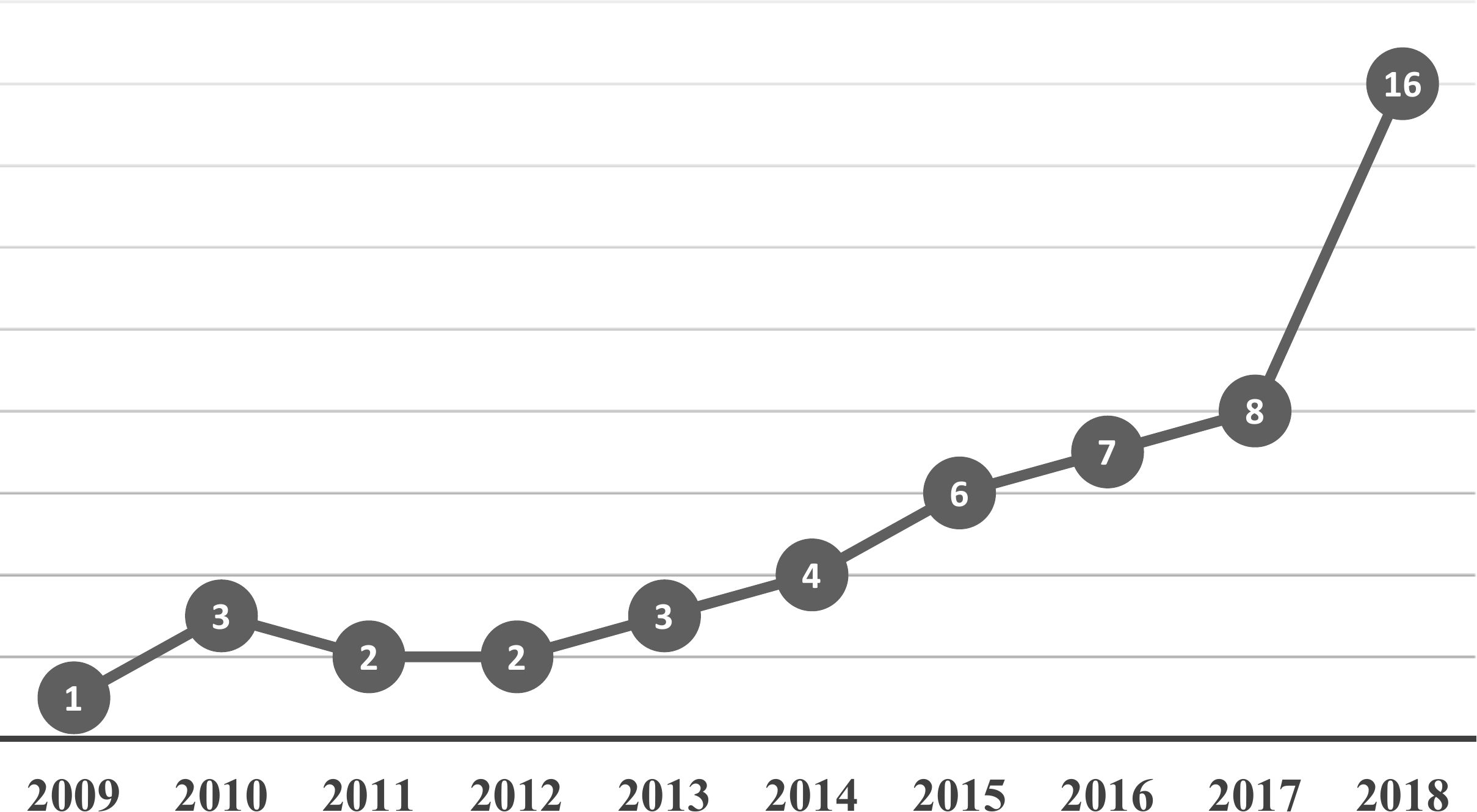}
  \caption{Publications per year}
  \label{fig:year}
\end{figure}

        We explained in section \ref{subsec:searchProcess} why we only included papers released since 2009. The collected papers show that the number of publications per year is stagnant between 2009 and 2012 at one to three papers. The years between 2012 and 2017 show a slight, but steady increase and in 2018 the amount of papers was doubled compared to the preceding year. The data collection was carried out in October 2018 and since then, a number of fitting papers have been released that haven't been reviewed in this study. This suggests that the trend towards increasing research in this field of study is nowhere near stopping. The exact numbers can be seen in Figure \ref{fig:year}. 
        At the beginning of the time period covered, in 2009, Song et al.~\cite{Song2009} proposes a pose-variant face recognition system based on BPNNs and Active Appearance  model (AAM). In 2018 Shen et al.~\cite{Shen2018} trained a CNN based on YOLO architecture to detect flames in video sequences.
        
    \subsubsection{Publication Form}
        28 Papers have been released as articles, 23 were published in conferences and a single one as part of a workshop. This means there is only a slight trend towards the publication in form of journal articles. Articles are generally considered to be of higher quality, due to the more self contained nature of the papers, whereas conference papers are designed to be openly discussed. 
        
    \subsubsection{Measured Quality}
        We measured the quality of each paper by applying the quality measurements explained in section \ref{subsec:QualityAssessment}. The asked yes/no questions were mostly aimed at aspects of the craftsmanship of the paper, such as its completeness, its quality of documentation or the inclusion of statistical reasoning. The scientific value of the publication had to be evaluated subjectively by the researchers by questioning the credibility, importance and magnitude of the research. Another, widely acknowledged parameter of quality is the amount of received citations of a paper. Since a lot of papers were published only recently, only a few have received more than single digit citation numbers. We measured the quality of the four most cited papers which are  Miljkovi\a' c et al.~\cite{Miljkovic2013}  with a measured quality of 75\%, Ghesu et al.~\cite{Ghesu2016} with 93\%, Pinto et al.~\cite{Pinto2013} with 73\%, and Franceschini~\cite{Franceschini2014} with 100\%. This shows a close correlation to our metrics and suggests that our measurement is sufficient for estimating the quality of the reviewed papers. 
        14 papers scored above 80\% on our index, indicating a very high quality of work. Notable examples are:
            (i) Franceschini~\cite{Franceschini2014}, which presented a very extensive collection of insect inspired robot developments over the last decades.
            (ii) Martins et al.~\cite{Martins2015}, where a very promising, new shape coding approach using proto object categorization was presented.
            (iii) Wen et al.~\cite{Wen2018}, where a novel object recognition system using radar spectograms as the environment-independent input data was presented.
        
        37 papers scored above 50\% and only two papers had quality ratings below 20\%. These numbers suggest that the majority of research is executed with sufficient rigor, showing the maturity of the field. A lot of authors aren't native English speakers, which led to varying quality of readability and comprehensibility. Therefore, a portion of the papers was very time consuming to read due to these factors. Since the quality of the language doesn't reflect the quality of the content, we have decided to give it only a low relevance in the measurement. While a large number of researchers give a longer introduction to the reader who is not familiar with the research field, a large proportion assume that the reader has domain knowledge that hinders readability for the untrained reader. 
        
  \subsection{Field of Research}
    
    The reviewed papers all stem from a wide range of research areas, all of which are more or less closely related to robotics. This means that the reached results were either directly developed for the use in a robotic environment or are still beneficial for robot development. In the first step, we categorized the areas exactly, resulting in only a few overlaps between separate studies, like robot navigation or object recognition. In the second step, we condensed these categories down into seven upper categories. 
    
    \subsubsection{General Development}
        Under the term fundamental research we grouped multiple papers which presented very low level approaches with no immediate real world usability. But by using their knowledge as a foundation, great future works are possible. Cui et al.~\cite{Cui2015} optimized restricted boltzmann machines (RBM) by restricting their data to sparse matrices, pushing the efficiency of RBMs greatly. Angeletti et al.~\cite{Angeletti2018} refines the foundations of image recognition by training neural networks with image background- instead of foreground data, enabling more flexibility when detecting actual objects.
        The next field of research is automated visual inspection. A system or object is monitored with a camera or a set of sensors, ensuring either its correct function or a lack of errors. Kadmin et al.~\cite{Kadmin2018} used radial basis function networks (RBF) in combination with a robotic arm to determine what class of consumer goods an object belongs to. Pei et al.~\cite{Pei2010} introduced a camera based system with location specific ANNs, which let a robot arm reach its target in a simulated environment, while being very efficient.  Qiu et al.~\cite{Qiu2012} used a camera to determine the magnitude of the vibrations of a component. Likewise, they were able to utilize a RBF to reduce these arbitrary vibrations, omitting the use of a cost- or space-intensive sensor. 
        Object tracking/detection summarizes the attempts at localizing, classifying, recognizing and tracking objects in single images or image sequences. Karayaneva and Hintea~\cite{Karayaneva2018} implemented different functions of the OpenCV computer vision library on a NAO robot, including the recognition of different colors and shapes for the means of child education. A more refined approach was presented by Kuremoto et al.~\cite{Kuremoto2016}, which correctly classified a number of hand gestures using self organizing maps (SOM) in unison with an asymmetric neighborhood function. Song et al.~\cite{Song2009} proposed an algorithm based on a back propagating NN (BPNN) that detects faces, even when they aren't pointing directly to the camera.
    
    \subsubsection{Robot Development}
        Visual servoing is the procedure of using collected image data directly to control a robotic system. Typically a specific feature will be selected and the robot heads towards it. Petković et al.~\cite{Petkovic2016} presents a visual servoing controller that utilizes fuzzy controls and neural networks to the task, improving it greatly. 
        Robot manipulation combines all activities, where a robot is used to alter a system, mainly by grasping and placing small objects. Haochen et al.~\cite{Haochen2017} described the training of a convolutional neural networks (CNN) to let a robot arm correctly localize and grasp three different types of circuit boards. Zhihong et al.~\cite{Zhihong2017} proposes a robot arm grasping system, which automatically detects and localizes items on a garbage conveyor belt with a fast recurrent CNN (Fast R-CNN). 
        Robot navigation has the goal of enabling robots to traverse known or unknown environments fully or partially autonomous. A very simple example, although only used as a test bed for a vastly more complex learning algorithm is the line following robot of Murali et al.~\cite{Murali2017}. Utilizing Q-learning with ANNs, the algorithm can use arbitrary sensory input data to do its tasks. For the line following, a live video feed from the front of the robot was used as the input. Prieto et al.~\cite{Prieto2012} presented research on swarm based robots. These monitor neighboring swarm robots and mimic their behaviour using automatic neural-based pattern classifiers (ANPAC). These were tested in simulations.
            
    \subsubsection{Quantitative Differences} 
        Of the 52 papers, the biggest field of research is object tracking/detection with 15 papers, followed by robot manipulation and robot navigation with 12 and eleven publications. Seven papers describe automatic visual inspection systems, five are fundamental researches and two center around visual servoing. There was no evidence observed, that the quality of the research correlated with specific research fields. All research areas had an equal distribution of high- and low quality papers.
        
  \subsection{Practicability}\label{sec:practicabel}
    
  We measured practicability by using a self defined numeric metric in the interval [1,10] giving a score to each paper's practical relevance in terms of how we evaluated their proposal. We gave a score of 1 to papers which were laying out foundations for other researches with no evaluation and a score of 10 when researchers tested their system or method for example on a real robotic system or mobile robot. 
  There are numerous factors on which practicability depends, naming: the used type of benchmark / experiment, the type of evaluation, the application area. Typical signs are furthermore the number of test objects,  included pictures of the real setup. On average, we gave papers a score of 5.1 points, which means that the included papers are not purely theoretical research, which we ruled out by usage of our search string.  This means that a portion of researchers have spent resources on non-simulation test scenarios. 

We conclude, that practicability does not correlates with quality because the chosen metrics evaluate the quality of writing rather than the proposal itself. For example~\cite{Pan2018} got a score of 37,5 (84\%) while being being not very practicable. On the other hand~\cite{Du2013} scored only 17 points but having a higher practical relevance (7). A similar case is~\cite{Chen2010} with high practicality but a  low quality score.  The most practical papers in our opinion  were~\cite{Wang2018} showing the system setup including a camera and a robotic arm and~\cite{Probst2018} with an evaluation of the proposal, which was conducted on real pig eyes.  
 On the other side of the spectrum,~\cite{Tang2017} experimented work piece recognition on very simple geometric shapes scoring low points.
    
 Why are some approaches practicable and others are not? We expect that this depends on the amount of time and resources spent on the project (a simulation is less expensive when lacking funding) and the current state-of-the-art achievable as well as the amount of improvement made through the study.  For example, Gerrard et al.~\cite{Gerrard2014} have carried out fundamental research, creating neural networks based on chemical reaction chains in cells, proving they can provide complex behaviour without complex neural systems.  
 For a proposal in an early stage it makes more sense to evaluate it on a small example and if its usefulness is proven than subsequent work can be implemented on real systems. 
 

  \subsection{How are the developments of the field structured?}
    
    A lot of papers describe the development of a new method, algorithm or system that is based on an already existing approach. If the base system is described in another paper from our review, interesting trees of development can be observed. Two such trees have been observed, one building on the widely popular CNN, the other one the related Regions with CNN (R-CNN). 
        \subsubsection{CNN}
\begin{figure}[h]
     \centering
     \includegraphics[width=0.8\columnwidth]{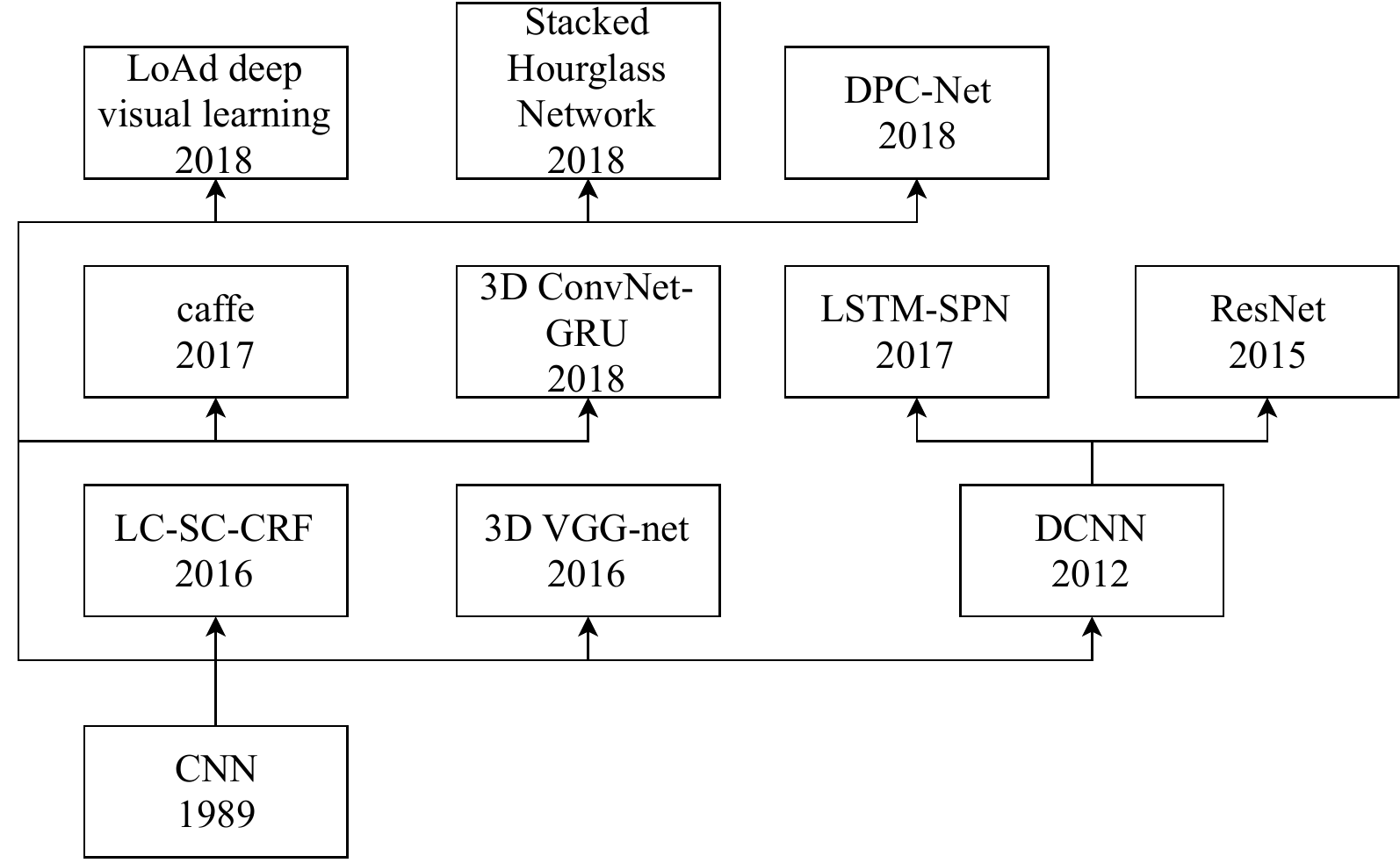}
     \caption{Technology hierarchy, starting from DNN}
     \label{fig:dnn_tree}
     \end{figure}

            A lot of papers found by our review are based on or are utilizing the concepts of CNNs. As Figure \ref{fig:dnn_tree} shows, a very wide graph can be constructed from these connections, which demonstrates the relevance of CNNs in the field of robot vision. This popularity stems from their impressive successes in many image classification benchmarks. The first such demonstration is described by Krizhevsky et al.~\cite{Krizhevsky2012}, where at the time groundbreaking top 1 error rates of 39.7\% were attained. Naturally, all subsequent ranking leaders were using CNNs. 
            CNNs are neural networks that utilize one or multiple convolutional layers followed by a pooling layer. Multiple such combinations are executed and finally finished by a fully connected layer. Through the unique representation of the learned weights, a lot less RAM is needed to extract the wanted features from images and respectable outcomes are achievable without the use of super computers. 
            There are multiple reasons that were stated on why researchers chose to use CNNs in their work or base it on them. The biggest one is, to no surprise, that they wanted their research their work on methods which are commonly viewed as the state of the art. Their research scope also usually isn't the development of CNNs. Such a paper is presented by Farazi et al.~\cite{Farazi2017}. Papers like Quin et al.~\cite{Qin2017} compare multiple methods and are ultimately choosing CNNs. Yeboah et al.~\cite{Yeboah2018} recognized CNNs as the state of the art and used that argumentation to try and enhance it. The most elaborate reasoning is presented by papers like Wen et al.~\cite{Wen2018}, where all the required features of an approach were listed and finally CNNs were chosen for the task. 
            Therefore, the ease of computation of complex computer vision can be stated as the main reason for the use of CNNs. After this fact was proved and functional examples and frameworks were made available, big parts of the research community simply relied on this insight. One may be tempted to critically question this development, since other, still unexplored approaches may be even more powerful in the specified tasks, but aren't being researched as widely because of the focus on CNNs. On the other hand, a lot of great discoveries and breakthroughs are made simply because a simple, reliable platform for computer vision research exists.

        \subsubsection{R-CNN}
\begin{figure}
     \centering
     \includegraphics[width=0.7\columnwidth]{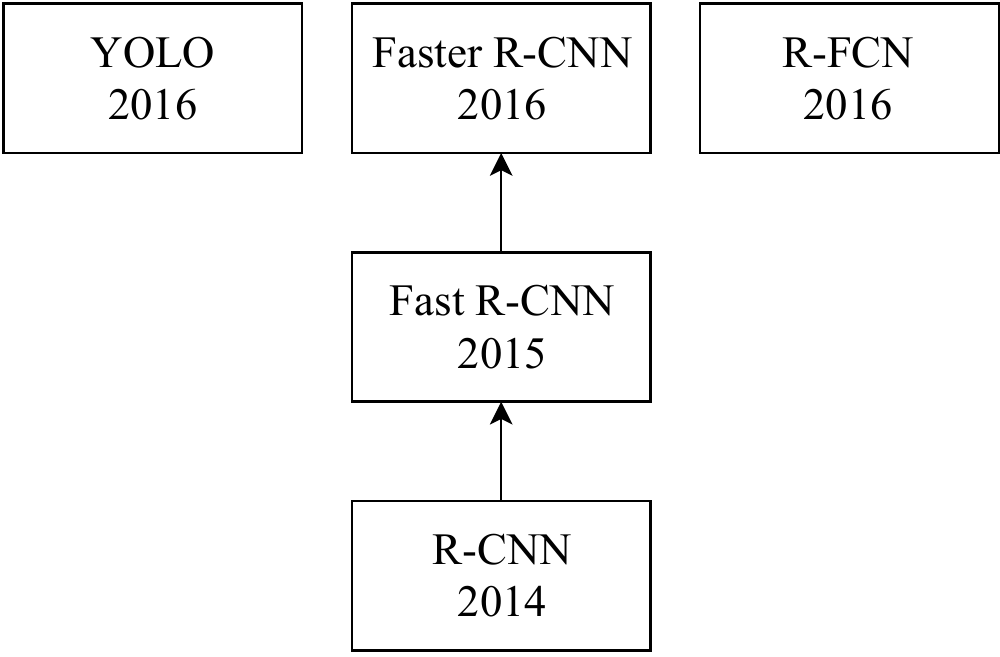}
     \caption{Technology hierarchy, starting from R-CNN}
     \label{fig:rcnn_tree}
     \end{figure}

            The second tree \ref{fig:rcnn_tree} shows, that all the major steps of the development of R-CNN and adjacent developments are present in our review scope. R-CNN was presented by Girshick et al.~\cite{Girshick2013} and utilized for a partial problem in Lee et al.~\cite{Lee2016}. R-CNN uses a selective search algorithm to divide a given image into regions, which are then analyzed by a CNN. Fast R-CNN was developed again by Girshick~\cite{girshick2015}. Instead of generating a lot of regions, the image is fed into the CNN to generate a single convolutional feature map, reducing the computation time immensely. Zhihong et al.~\cite{Zhihong2017} use Fast R-CNN to recognize and localize objects on a garbage conveyor belt, enabling a robot to grasp them in real time. The last subsequent step is Faster R-CNN by Ren et al.~\cite{Ren2015}, where a separate Network predicts the region proposals. Lee et al.~\cite{Lee2016} integrates Faster R-CNN into multiple CNN architectures, concluding that ResNet provides the highest precision of the tested models. Fu et al.~\cite{Fu2018} uses Faster R-CNN in combination with a Zeiler and Fergus network (ZFNet) to detect the exact count and position of kiwifruits in photos taken on the field.
            Another, comparable approach is the You Only Look Once (YOLO) method proposed by Redmon et al.~\cite{redmon2015}. Instead of dividing the image into separate parts, a single convolutional network generates class probabilities for predicted bounding boxes. This makes the model very fast, whilst accuracy especially in small details is sacrificed. The model is used by Llopart et al.~\cite{Llopart2017} to detect doors and door handles for an autonomous robot navigation system. The system described by Wang et al.~\cite{Wang2018} needs to localize, classify and finally sort many small objects in a short amount of time. Faster R-CNN and YOLO are both considered but ultimately rejected in favor of Region-based full convolutional networks (R-FCN) described by Dai et al.~\cite{Dai2016}. This network has the same approach as YOLO by using position sensitive score maps to classify whole images. Ultimately it provides a better balance between accuracy and speed than YOLO. Interestingly, no other reviewed paper used R-FCNs, which may be due to its low age.
                
        \subsubsection{Lessons learned}
            Both observed trees show interesting properties of popular robot vision approaches. In the case of CNNs, a first come, first served mindset is visible. The first effective, functional approach to machine vision is used the most as a base for other developments. R-CNNs, which are also based on CNNs, show a different course of development, with a steady stream of improvements and the subsequent overthrow by other, more performant developments.


  \subsection{Research Questions}
  In the next section we discuss our findings relating to the research questions.
    
    \subsubsection{
   Studies on which image processing techniques result in the most practical applications}
We investigate this issue by using a numeric metric giving a score [1,10] to each paper on its practical relevance. A lower score indicates that only a theoretical foundation was proposed or the evaluation was conducted in a simulation. On the other side, a high score means that the experiment was evaluated on a real robotic system, preferably in real world conditions. The numbers in between represent gradations between these two extremes. The most practical papers included~\cite{Probst2018},~\cite{Wang2018},~\cite{Wen2018},~\cite{Fu2018} and~\cite{Pinto2013}, where researchers used robotics systems in the evaluation of their proposals. 
The method of evaluation was already discussed  in subsection \ref{sec:practicabel}. We took the papers we evaluated as practical and observed the techniques used. For practicality often a low enough computation speed is needed.  Which of the  techniques  produce the most practical applications depends highly  on the type of problem solved. In object recognition 2D-CNN architectures are used by relying on  architectures build upon ImageNet~\cite{deng2009imagenet} classifiers such as AlexNet, ResNet and GoogLeNet. These have had great success on the popular ILSVRC\footnote{\url{http://image-net.org/challenges/LSVRC/}} challenge starting with the breakthrough by AlexNet~\cite{Krizhevsky2012} which used GPU computation in the training step. 
CNNs are used in vision applications because this type of data nearly always has spatial relationships between related objects in the image. 
The computation complexity of high resolution images is reduced by down sampling and using sliding windows scanning the whole image and selecting a region of interest (RoI). 
 In object localization (by which we mean the detection of objects in an image and segmentation of background) CNNs as well as improvements building upon it like R-CNN are being used for example in~\cite{Llopart2017} where YOLO~\cite{redmon2015} model (proposed in 2015) is used to identify handles of doors and estimate its pose for a grasping application tested on a mobile robot. There is an even faster, lightweight variant called Faster YOLO achieving higher processing speed. ZFNet~\cite{DBLP:journals/corr/ZeilerF13} - a faster R-CNN variant, is used in~\cite{Fu2018} to detect multiple kiwifruits from images in clustered scenes.
 When taking video sequences as as input, typical problems are the classification of actions or motion planning. Here DCNN networks are used for example the VGG~\cite{Simonyan14c} architecture~\cite{Yao2018}), as well as Self Organizing Maps network which in~\cite{Najmaei2011} receives information about the human's location and pose in a robot work space based on pressure activated notes in a safety mat.


      \subsubsection{
    Which problems are typically solved by the studies in this research field and which ones are still unsolved}
    
     
     Most reviewed papers are trying to solve one or more discrete problems which were imposed by their previous research. As a result, multiple classes of papers can be found. Approaches showcase the foundations and a short proof of work for a novel or improved, small subarea of a research field. The papers Enikov and Escareno~\cite{Enikov2015}, Olaque et al.~\cite{Olague2018} and Pinto et al.~\cite{Pinto2013} were classified as approaches. Methods are more rigorous, as they have the aim of presenting a ready to deploy method, tested and validated for the intended use. Chen et al.~\cite{Chen2014}, Peretroukhin and Kelly~\cite{Peretroukhin2018}, Shirzadeh et al.~\cite{Shirzadeh2015} and Rupprecht et al.~\cite{Rupprecht2016} present methods. Systems, frameworks and architectures present a whole environment, in which multiple approaches and methods are combined and their interaction with each other and the outside world is designed. Examples are Lin et al.~\cite{Lin2018}, Sanders et al.~\cite{Sanders2010} and Calli et al.~\cite{Calli2018}. The class algorithm presents the conception and testing of a single algorithm, as demonstrated by Li et al.~\cite{Li2014}. Other classes like reviews, comparisons and implementations are trying to solve different, tertiary problems and shall not be discussed here. 
     Each of these classes has a different tendency to what problems are trying to be solved and which problems arise during the research. Most researchers define a narrow range, in which their solution is settled and therefore works best. The limitations imposed by such a restriction were also considered problems.
     
     \begin{itemize}
         \item Approaches mostly solved non-complex object recognition tasks. Shaker and ElHelw~\cite{Shaker2017} present an easier to train OCR model, Shen et al.~\cite{Shen2018} trained a CNN to detect flames in an image sequence and Qiu et al.~\cite{Qiu2012} use a RBF neural network to reduce vibrations in a flexible manipulator. The common flaw of these approaches is their very narrow field of work. Each is tested only on a small sample size or in a simple environment, so a lack in generalization abilities can be assumed. Of course, as stated before, this is the intended area of approaches.
         
         \item The majority of methods try to solve object classification and localization problems like Chen et al.~\cite{Chen2014} and Llopart et al.~\cite{Llopart2017}, which are both detection doors in rooms. The methods are tested in appropriately complex environments, reducing the simplification error seen in approaches. Errors are now harder to be overseen and can therefore be addressed easier. The most reported flaws are the accumulation of accuracy errors. 
         
         \item Papers on systems, frameworks and architectures are presenting whole environments and are therefore able to address a wide range of problems. All kinds of robot vision field of researches, like action recognition, motion planning and gesture recognition are dealt with. Occlusion, where parts of an object are covered by the environment or other objects, is a big problem, as described by Fu et al.~\cite{Fu2018} and Wang et al.~\cite{Wang2018}. The amount of training data needed for sufficient recognition abilities is problematic for papers like Farazi et al.~\cite{Farazi2017}. Many papers state the goal of achieving real time results, but have to ultimately cut corners on the accuracy to reach the desired times. Insufficient sensors are stated as problems by papers like Najmaei and Kermani~\cite{Najmaei2011} and Quin et al.~\cite{Qin2017}. 
     \end{itemize}
     
     A huge problem researchers in all classes addressed is the variability of lighting in scenes. It changes the environment of object recognition algorithms drastically and increases the amount of training data needed. Therefore, it already is its own field of research. Wen et al.~\cite{Wen2018} approaches the problem by using radar spectograms as a light invariant form of imaging. Angeletti et al.~\cite{Angeletti2018} tries to train domain invariant features to a CNN to make it recognize object features which don't change with lighting.

   \subsection{Threats to validity}
 
Identified threats to the validity of our study and possible causes are:

\paragraph{ External  validity} Because the selected research field in this study is really broad, we assume that it represents a cross-section of the available research.  However,  our study may be restricted by not covering the whole research field. The coverage depends on the chosen search string and is hard to measure. However a more general search string would imply a more time consuming process, including more potentially relevant papers.
\paragraph{ Internal and construct validity} The internal validity may be threatened by the fact that only 2 researchers extracted the data, which could potentially lead to human errors and biased results. However in critical situations, the results were double checked by a third researcher.

\section{Conclusion} \label{sec:conclusion}
Lastly, considering
the results from this systematic literature review (SLR), potential future research directions are suggested.
~As robustness and computation time are two key-component for real time applications, we assume that the research field covered in this study will continue to find possible improvements in these areas just like the numerous approaches inspired by the  ILSVRC 2012 winner - AlexNet.   As occlusion and lighting variance pose challenges to the field, we further expect that these will be addressed for example. A portion of the reviewed work was in an earlier stage of research, so subsequent papers demonstrating the capabilities on more than simple experiments is anticipated. 
In conclusion, the problem of finding the balance
between efficiency and accuracy
is a dilemma in robot vision as much as any other
research area. In our future work, we will utilize the knowledge acquired in this study to solve an object recognition problem in an cyber physical systems environment in real time. 

\clearpage
\appendices
\begin{table}[htbp]
  \centering
    \begin{tabular}{|p{7cm}|r|}
    \toprule
    \rowcolor[rgb]{ .867,  .494,  .42} \textbf{Set Up} & Value \\
    \midrule
    \rowcolor[rgb]{ .953,  .953,  .953} Are the aims clearly stated? & \cellcolor[rgb]{ .835,  .651,  .741}3 \\
    \midrule
    \rowcolor[rgb]{ .953,  .953,  .953} Are the data collection methods adequately described? & \cellcolor[rgb]{ .835,  .651,  .741}3 \\
    \midrule
    \rowcolor[rgb]{ .953,  .953,  .953} Are the measures used in the study the most relevant ones for answering the research questions? & \cellcolor[rgb]{ .835,  .651,  .741}3 \\
    \midrule
    \rowcolor[rgb]{ .953,  .953,  .953} Are statistical methods used? & \cellcolor[rgb]{ .835,  .651,  .741}3 \\
    \midrule
    \rowcolor[rgb]{ .867,  .494,  .42} \textbf{Execution} &  \\
    \midrule
    \rowcolor[rgb]{ .953,  .953,  .953} Was the data collection carried out well? & \cellcolor[rgb]{ .835,  .651,  .741}3 \\
    \midrule
    \rowcolor[rgb]{ .953,  .953,  .953} Are scoring systems described? & \cellcolor[rgb]{ .835,  .651,  .741}3 \\
    \midrule
    \rowcolor[rgb]{ .953,  .953,  .953} Has the approach to, and formulation of, analysis been conveyed well? & \cellcolor[rgb]{ .835,  .651,  .741}3 \\
    \midrule
    \rowcolor[rgb]{ .953,  .953,  .953} Is the reporting clear and coherent? & \cellcolor[rgb]{ 1,  .949,  .8}1 \\
    \midrule
    \rowcolor[rgb]{ .953,  .953,  .953} Has the research process been documented adequately? & \cellcolor[rgb]{ .835,  .651,  .741}3 \\
    \midrule
    \rowcolor[rgb]{ .867,  .494,  .42} \textbf{Evaluation} &  \\
    \midrule
    \rowcolor[rgb]{ .953,  .953,  .953} Was statistical significance assessed? & \cellcolor[rgb]{ .835,  .651,  .741}3 \\
    \midrule
    \rowcolor[rgb]{ .953,  .953,  .953} Are all study questions answered? & \cellcolor[rgb]{ .635,  .769,  .788}4 \\
    \midrule
    \rowcolor[rgb]{ .953,  .953,  .953} Are the findings credible? & \cellcolor[rgb]{ .635,  .769,  .788}4 \\
    \midrule
    \rowcolor[rgb]{ .953,  .953,  .953} If credible, are they important? & \cellcolor[rgb]{ .635,  .769,  .788}4 \\
    \midrule
    \rowcolor[rgb]{ .953,  .953,  .953} Has knowledge or understanding been extended by the research? & \cellcolor[rgb]{ .635,  .769,  .788}4 \\
    \bottomrule
    \end{tabular}%
    \caption{Table of used quality metrics with each assigned numerical value.}
    \label{tab:qualityMeas}%
\end{table}%

      \begin{figure*}
\centering
\begin{lstlisting}[linewidth=18cm, breaklines=true,  keywordstyle=\bfseries,showstringspaces=false, morekeywords={AND, OR, NOT}, resetmargins=true,   keepspaces=true,]
(((("object recognition" OR "computer vision" OR "image processing" OR "pattern recognition" OR "pedestrian detection" ) AND ("cyber physical systems" OR "robotics" OR "cobot" OR "collaborative robot" OR "robot application" OR "robot arm")  AND ("application" OR "methodology" OR "method" OR "framework") AND "model*" AND "control*" AND ("deep-learning" OR "deep learn*" OR "neural network")  NOT "remote" NOT "deforma*" NOT "theory")))
\end{lstlisting}
\caption{Used search string in database \label{fig:searchstringIEEE}}
\end{figure*}


\begin{figure*}
\centering
\begin{lstlisting}[linewidth=18cm, breaklines=true,   keywordstyle=\bfseries,showstringspaces=false, morekeywords={AND, OR, NOT}, resetmargins=true,   keepspaces=true,]
(("object recognition" "computer vision" "image processing" "pattern recognition" "pedestrian detection") AND ("cyber physical systems" "robotics" "cobot" "collaborative robot" "robot application" "robot arm") AND ("application" "method" "methodology" "framework") AND ("model*") AND ("control*") AND ("deep learn*" "deep-learn*" "neural network*") -"remote" -"deforma*" -"theory")
filter: {"publicationYear":{ "gte":2009 }}
{owners.owner=HOSTED}
\end{lstlisting}
\caption{Used search string in database ACM\label{fig:searchstringACM}}
\end{figure*}

\begin{table*}[htbp]
  \centering
    \begin{tabular}{p{10.5cm}p{0.75cm}rp{1.5cm}p{1.5cm}}
    \toprule
    \textcolor[rgb]{ 0,  0,  0}{\textbf{Title}} & \multicolumn{1}{l}{\textcolor[rgb]{ 0,  0,  0}{\textbf{Year}}} & \textcolor[rgb]{ 0,  0,  0}{\textbf{\# cites}} & \textbf{quality score (max: 44)} & \textbf{practical value (max: 10)} \\
    \rowcolor[rgb]{ .859,  .859,  .859} Neural network Reinforcement Learning for visual control of robot manipulators \cite{Miljkovic2013} & \multicolumn{1}{l}{2013} & 61 & 33 & 5 \\
    Marginal Space Deep Learning: Efficient Architecture for Volumetric Image Parsing \cite{Ghesu2016} & \multicolumn{1}{l}{2016} & 56 & 41  &6\\
    \rowcolor[rgb]{ .859,  .859,  .859} Small brains, smart machines: From fly vision to robot vision and back again \cite{Franceschini2014} & \multicolumn{1}{l}{2014} & 36 & 44 & / \\
    Object recognition using laser range finder and machine learning techniques \cite{Pinto2013} & \multicolumn{1}{l}{2013} & 36 & 32 & 8\\
    \rowcolor[rgb]{ .859,  .859,  .859} Applications of artificial intelligence in safe human-robot interactions \cite{Najmaei2011} & \multicolumn{1}{l}{2011} & 27 & 34 & 7  \\
    Input Displacement Neuro-fuzzy Control and Object Recognition by Compliant Multi-fingered Passively Adaptive Robotic Gripper \cite{Petkovic2016} & \multicolumn{1}{l}{2016} & 25 & 31 & 4 \\
    \rowcolor[rgb]{ .859,  .859,  .859} A robotic welding system using image processing techniques and a CAD model to provide information to a multi-intelligent decision module \cite{Sanders2010} & \multicolumn{1}{l}{2010} & 24 & 37 & 6\\
    Door recognition and deep learning algorithm for visual based robot navigation \cite{Chen2014}& \multicolumn{1}{l}{2014} & 21 & 23 &5 \\
    \rowcolor[rgb]{ .859,  .859,  .859} Action Recognition Based on Efficient Deep Feature Learning in the Spatio-Temporal Domain \cite{Husain2016} & \multicolumn{1}{l}{2016} & 16 & 20 & 7\\
    Intelligent control based on wavelet decomposition and neural network for predicting of human trajectories with a novel vision-based robotic \cite{Soyguder2011} & \multicolumn{1}{l}{2011} & 15 & 23.5 & 4 \\
    \rowcolor[rgb]{ .859,  .859,  .859} An indirect adaptive neural control of a visual-based quadrotor robot for pursuing a moving target \cite{Shirzadeh2015} & \multicolumn{1}{l}{2015} & 11 & 22 & 3 \\
    Sensor Substitution for Video-based Action Recognition  \cite{Rupprecht2016} & 2016  & 10 & 37 & 5 \\
    \bottomrule
    \end{tabular}%
    \caption{Most frequently cited papers by year}
  \label{tab:mostcited} 
\end{table*}%

      \begin{figure}
\centering
  \begin{subfigure}[t]{0.48\linewidth}   
  \centering
    \includegraphics[height=5cm]{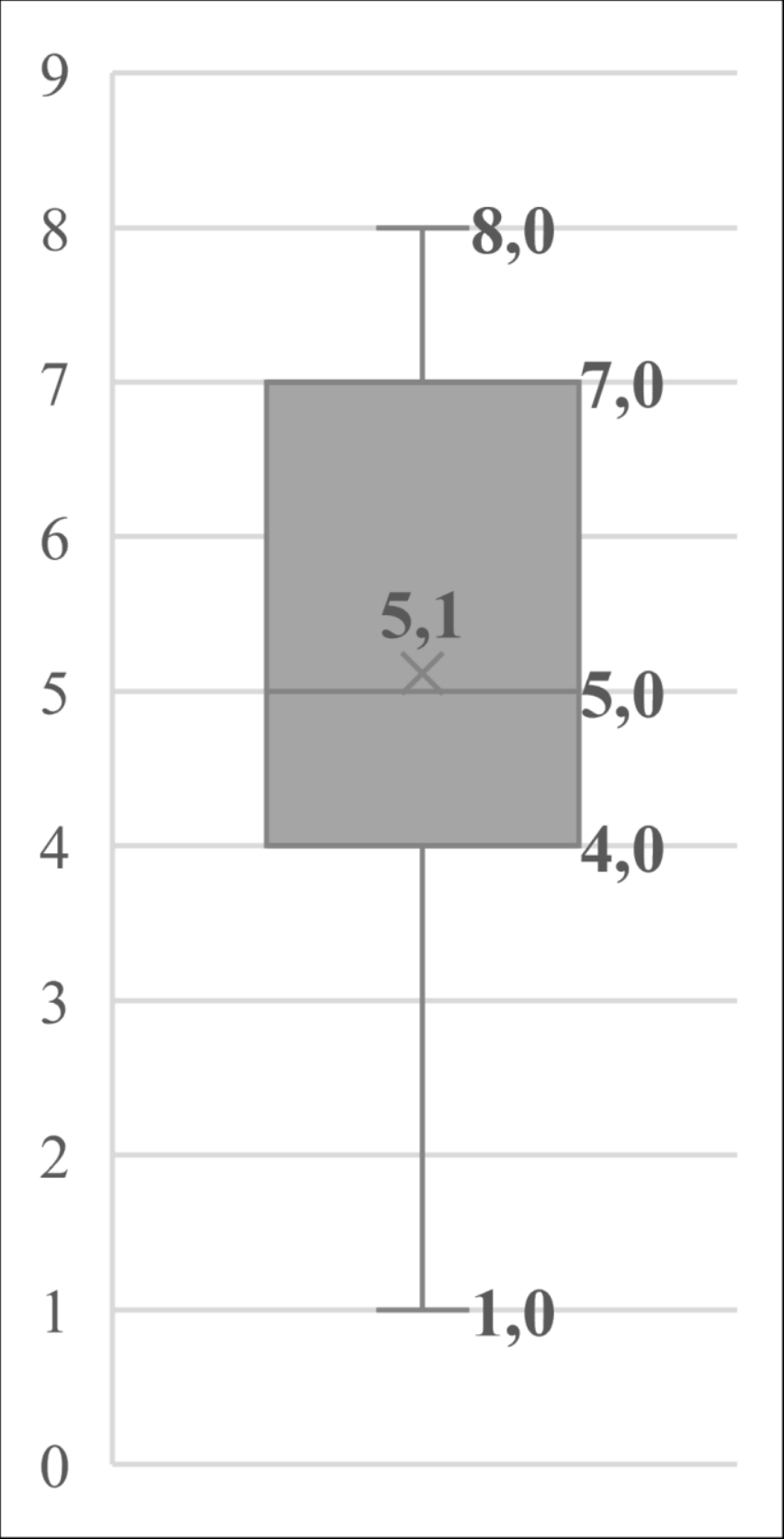}
    \caption{Distribution of practicality values}
    \label{fig:1}
  \end{subfigure}
  \begin{subfigure}[t]{0.48\linewidth}  
  \centering 
    \includegraphics[height=5cm]{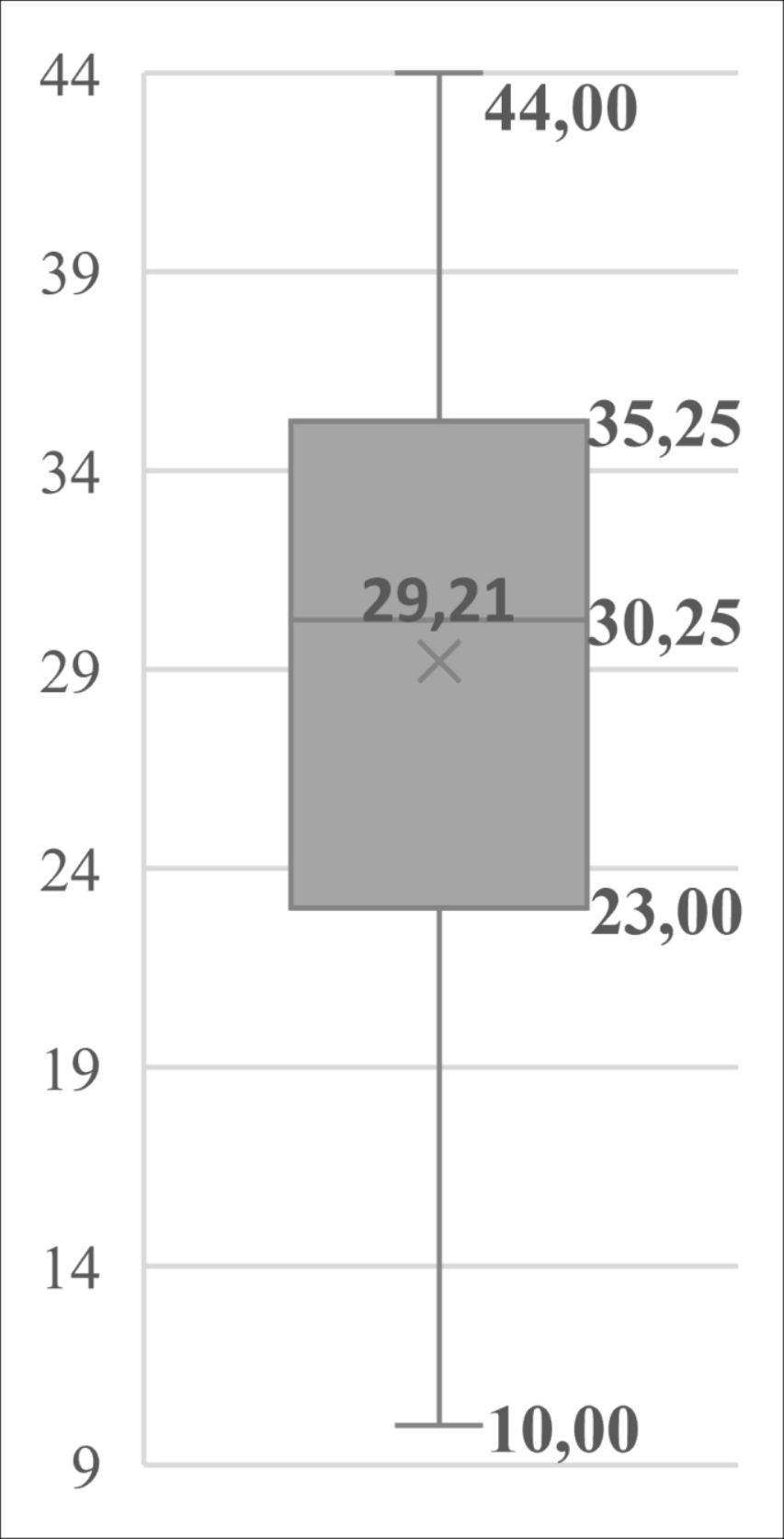}
    \caption{Distribution of quality values}
    \label{fig:2}
  \end{subfigure}
  \caption{Max, Q3, Q2, Mean, Q1, Min}
\end{figure}

\end{document}